# Joint Binary Neural Network for Multi-label Learning with Applications to Emotion Classification

Huihui He, Rui Xia*

School of Computer Science and Engineering,
Nanjing University of Science and Technology
hehuihui1994@gmail.com, rxia@njust.edu.cn

## Abstract

Recently the deep learning techniques have achieved success in multi-label classification due to its automatic representation learning ability and the end-to-end learning framework. Existing deep neural networks in multi-label classification can be divided into two kinds: binary relevance neural network (BRNN) and threshold dependent neural network (TDNN). However, the former needs to train a set of isolate binary networks which ignore dependencies between labels and have heavy computational load, while the latter needs an additional threshold function mechanism to transform the multi-class probabilities to multi-label outputs. In this paper, we propose a joint binary neural network (JBNN), to address these shortcomings. In JBNN, the representation of the text is fed to a set of logistic functions instead of a softmax function, and the multiple binary classifications are carried out synchronously in one neural network framework. Moreover, the relations between labels are captured via training on a joint binary cross entropy (JBCE) loss. To better meet multi-label emotion classification, we further proposed to incorporate the prior label relations into the JBCE loss. The experimental results on the benchmark dataset show that our model performs significantly better than the state-of-the-art multi-label emotion classification methods, in both classification performance and computational efficiency.

## 1 Introduction

Multi-label emotion classification, is a sub-task of the text emotion classification, which aims at identifying the coexisting emotions (such as joy, anger and anxiety, etc.) expressed in the text, has gained much attention due to its wide potential applications. Taking the following sentence

Example 1: "*Feeling the warm of her hand and the attachment she hold to me, I couldn't afford to move even a little, fearing I may lost her hand*" for instance, the co-existing emotions expressed in it contain *joy*, *love*, and *anxiety*.

Traditional multi-label emotion classification methods normally utilize a two-step strategy, which first requires to develop a set of hand-crafted expert features (such as bag-of-words, linguistic features, emotion lexicons, etc.), and then makes use of multi-label learning algorithms [Xu *et al.*, 2012; Li *et al.*, 2015; Wang and Pal, 2015; Zhou *et al.*, 2016; Yan and Turtle, 2016] for multi-label classification. However, the work of feature engineering is labor-intensive and time-consuming, and the system performance highly depends on the quality of the manually designed feature set. In recent years, deep neural networks are of growing attention due to their capacity of automatically learn the internal representations of the raw data and integrating feature representation learning and classification into one end-to-end framework.

Existing deep learning methods in multi-label classification can be roughly divided into two categories:

- Binary relevance neural network (BRNN), which constructs an independent binary neural network for each label, where multi-label classification is considered as a set of isolate binary classification tasks and the prediction of the label set is composed of independent predictions for individual labels.

- Threshold dependent neural network (TDNN), which normally constructs one neural network to yield the probabilities for all labels via a softmax function, where the probabilities sum up to one. Then, an additional threshold mechanism (e.g., the calibrated label ranking algorithm) is further needed to transform the multi-class probabilities to multi-label outputs.

The structure of BRNN and TDNN are shown in Figure 1 (a) and (b), respectively.

However, both kinds of methods have their shortcomings. The former one, BRNN, usually known in the literature as binary relevance (BR) transformation [Spyromitros *et al.*, 2008], not only ignores dependencies between labels, but also consumes much more resources due to the need of training a unique classifier and make prediction for each label. The latter one, TDNN, although has only one neural network, can only yield the category probabilities of all class labels. Instead, it needs an additional threshold function mechanism to transform the category probabilities to multi-label outputs.

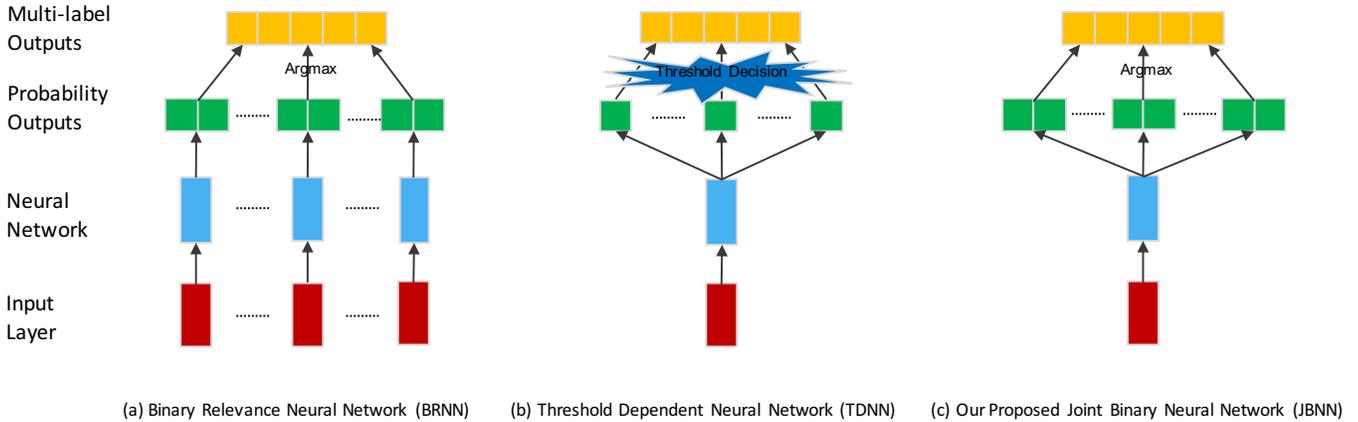

Figure 1: Different ways of constructing neural networks for multi-label classification.

However, building an effective threshold function is also full of challenges for multi-label learning [Zhang and Zhou, 2006; Read and Perez-Cruz, 2014; Nam *et al.*, 2014; Xu *et al.*, 2017; Lenc and Král, 2017].

In this paper, we propose a simple joint binary neural network (JBNN), to address these two problems. We display the structure of JBNN in Figure 1 (c). As can be seen, in JBNN, the bottom layers of the network are similar to that in TND-D. Specifically, we employ a Bidirectional Long Short-Term Memory (Bi-LSTM) structure to model the sentence. The attention mechanism is also constructed to get the sentence representation. After that, instead of a softmax function used in TDNN, we feed the representation of a sentence to multiple logistic functions to yield a set of binary probabilities. That is, for each input sentence, we conduct multiple binary classifications synchronously in one neural network framework. Different from BRNN, the word embedding, LSTMs, and the sentence representation are shared among the multiple classification components in the network. Moreover, the relations between labels are captured based on a joint binary learning loss. Finally, we convert the multi-variate Bernoulli distributions into multi-label outputs, the same as BRNN. The JBNN model is trained based on a joint binary cross entropy (JBCE) loss. To better meet the multi-label emotion classification task, we further proposed to incorporate the prior label relations into the JBCE loss. We evaluate our JBNN model on the widely-used multi-label emotion classification dataset Ren-CECps [Quan and Ren, 2010]. We compare our model with both traditional methods and neural networks. The experimental results show that:

- Our JBNN model performs much better than the state-of-the-art traditional multi-label emotion classification methods proposed in recent years;

- In comparison with the BRNN and TDNN systems, our JBNN model also shows the priority, in both classification performance and computational efficiency.

## 2 Related Work
### 2.1 Multi-label Learning

We first briefly review the traditional multi-label learning work and then lay the emphasis on neural network based methods.

The traditional multi-label classification methods can be classified into two main types: problem transformation methods and algorithm adaptation methods [Zhang and Zhou, 2014]. Problem transformation methods are the most direct way to deal with multi-label classification. They transform a multi-label classification problem into single-label problems, such as several binary problems [Spyromitros *et al.*, 2008; Read *et al.*, 2009; Sun and Kudo, 2015; Zhang *et al.*, 2015; Zhou *et al.*, 2017], multi-class problems [Read, 2008; Tsoumakas and Vlahavas, 2007] or label ranking problems [Hüllermeier *et al.*, 2008; Fürnkranz *et al.*, 2008]. Algorithm adaptation methods extend existing single-label classification algorithms to deal with multi-label classification [Cheng and Hüllermeier, 2009; Zhang and Zhang, 2010; Huang *et al.*, 2012; Zhang and Zhou, 2014].

In recent years, neural network (NN) approaches are of growing attention. In the field of multi-label learning, many neural network approaches have emerged in recent years.

As we have mentioned, neural networks can be constructed for multi-label classification in two different ways, *i.e.*, BRNN and TDNN. Feng *et al.* [2017] took the BRNN structure by transforming the original multi-label dataset into a single-label dataset to train a number of independent binary convolutional networks (CNNs). Most of the works belong to the latter. Zhang *et al.* [2006] proposed a BP-MLL model, adapted from a 3-layer forward neural network, to take dependencies between labels into account by requiring the labels belonging to an instance should be ranked higher than those not belonging to that instance. Finally, a threshold decision was used to get the multi-label output. Nam *et al.* [2014] replaced BP-MLL's pairwise ranking loss with binary cross entropy and use the TF-IDF representation of documents as network input. Read *et al.* [2014] used restricted Boltzmann machines to develop better feature representations. Kurata *et*

*al.* [2016] leveraged the co-occurrence of labels in the multi-label learning based on a convolutional network. Lenc *et al.* [2017] used standard feed-forward networks and popular convolutional networks (CNNs) with thresholding to obtain the final classification result. However, these NN-based multi-label classification algorithms mostly need an extra threshold mechanism to generate the multi-label outputs.

## 2.2 Multi-label Emotion Classification

We now focus on reviewing the approaches that were specific for multi-label classification. Xu *et al.* [2012] proposed a coarse-to-fine strategy for multi-label emotion classification. They dealt with the data sparseness problem by incorporating the transfer probabilities from the neighboring sentences to refine the emotion categories. Li *et al.* [2015] recast multi-label emotion classification as a factor graph inferring problem in which the label and context dependence are modeled as various factor functions. Yan *et al.* [2016] built a separate binary classifier for each emotion category to detect if an emotion category were present or absent in a tweet with traditional unigram features.

The deep learning techniques have also been employed in emotion classification. Zhou *et al.* [2016] proposed an emotion distribution learning (EDL) method, which first used recursive auto-encoders (RAEs) to extract features and then conducted multi-label emotion classification by incorporating the label relations into the cost function. Different from EDL where text representation and multi-label classification are separated, our JBNN model provides an end-to-end learning framework by integrating representation learning and multi-label classification in one neural network.

Wang *et al.* [2016] employed the TDNN framework by constructing a convolutional neural network (CNN) for multi-class classification at first, and then using the calibrated label ranking (CLR) algorithm to get multi-label outputs. Since the CLR algorithm depends on the binary classification score for each label, the BRNN framework was also adopted by constructing binary CNN classifier for each label. By contrast, our JBNN is a joint one-step learning method which neither needs additional effective threshold function such as TDNN nor requires training a set of isolate binary classifiers such as BRNN.

## 3 Model

In this section, we describe the proposed JBNN model in detail.

### 3.1 Joint Binary Neural Network

A Bi-LSTM structure is first employed to model the sentence. On the basis of Bi-LSTM, we propose our Joint Binary Neural Network (JBNN) for multi-label emotion classification. The structure of JBNN is shown in Figure 2.

Before going into the details of JBNN, we first introduce some notations. Suppose $E=\{e_1, e_2, \ldots, e_m\}$ is a finite domain of possible emotion labels. Formally, multi-label emotion classification may be defined as follows: giving the dataset $D=\{(x^{(k)}, y^{(k)}) | k=1, \ldots, N\}$ where $N$ is the number of examples in the $D$. Each example is associated with a subset of $E$ and this subset is described as an $m$-dimensional vector $y^{(k)}=\{y_1, y_2, \ldots, y_m\}$ where $y_j^{(k)}=1$ only if sentence $x^{(k)}$ has emotion label $e_j$, and $y_j^{(k)}=0$ otherwise. Given $D$, the goal is to learn a multi-label classifier that predicts the label vector for a given example. An example is a sentence in emotion classification.

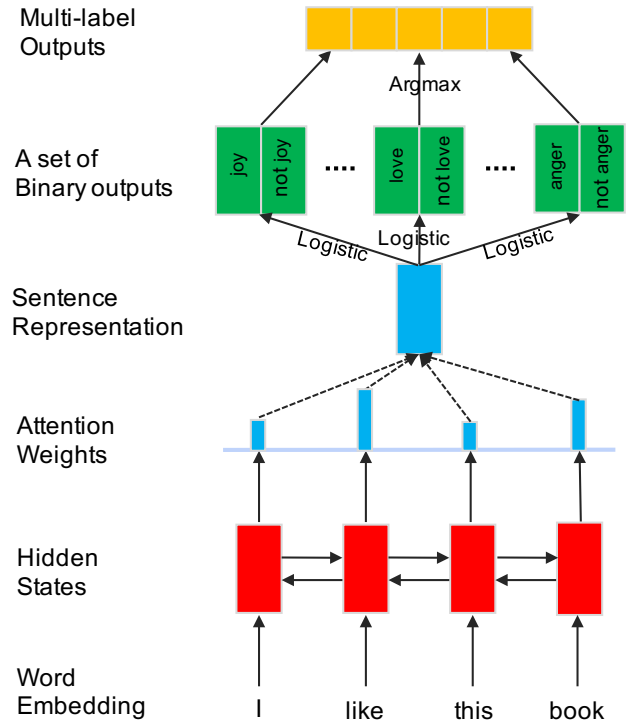

Figure 2: Overview of the Joint Binary Neural Network.

As shown in Figure 2, in JBNN, each word is represented as a low dimensional, continuous and real-valued vector, also known as word embedding [Bengio *et al.*, 2003; Mikolov *et al.*, 2013]. All the word vectors are stacked in a word embedding matrix $L_w \in R^{d \times |V|}$, where $d$ is the dimension of word vector and $|V|$ is vocabulary size. After we feed word embedding to Bi-LSTM, we can get hidden states $[h_1, h_2, \ldots, h_n]$ for a sentence as the initial representation.

Since not all words contribute equally to the representation of the sentence, we adopt the attention mechanism [Bahdanau *et al.*, 2014; Yang *et al.*, 2016] to extract such words that are important to the meaning of the sentence. Assume $h_t$ is the hidden states outputted in Bi-LSTM. We use an attention function to aggregate the initial representation of the words to form the attention vector $v$, also called sentence representation. Firstly, we use

$$u_t = \tanh(w_1 h_t + b_1), \quad (1)$$

as a score function to calculate the importance of $h_t$ in the sentence, where $w_1$ and $b_1$ are weight matrix and bias respectively. Then we get a normalized importance weight $\alpha_t$ for

the sentence through a softmax function:

$$\alpha_t = \frac{\exp(u_t^T u_1)}{\sum_t \exp(u_t^T u_1)}. \quad (2)$$

After computing the word attention weights, we can get the final representation $v$ for the sentence using equation:

$$v = \sum_t \alpha_t h_t. \quad (3)$$

After getting the sentence representation $v$, traditional Bi-LSTM based classification model normally feed $v$ into a softmax function to yield multi-class probabilities for multi-class classification. Our JBNN model differs from the standard model in that, we feed the feature vector $v$ to $C$ logistic functions, instead of a softmax function, to predict a set of binary probabilities $\{p(y_j = 1 \mid x), j = 1, \ldots, C\}$.

$$p(y_j = 1 \mid x) = p_j = \frac{1}{1 + e^{w_{lj}v + b_{lj}}}, \quad (4)$$

$$p(y_j = 0 \mid x) = 1 - p_j, \quad (5)$$

where $w_{li}$ and $b_{li}$ are the parameters in $j$-th logistic component.

Each component will receive a binary probabilities which determines whether this label is True or False in the current instance (*i.e.*, whether the label belongs to the instance):

$$\hat{y}_j = \arg\max_{y_j} p(y_j \mid x). \quad (6)$$

At last, we concatenate $\hat{y}_j$ to form the final predictions $\hat{y} = [\hat{y}_1, \ldots, \hat{y}_C]$.

### 3.2 Joint Binary Cross Entropy Loss with Label Relation Prior

The JBNN model can be trained in a supervised manner by minimizing the following Joint Binary Cross Entropy (JBCE) loss function:

$$L = -\sum_j^C \left( y_j \log p_j + (1 - y_j) \log(1 - p_j) \right) + \lambda ||\theta||^2, \quad (7)$$

where $\lambda$ is the weight for $L_2$-regularization, and $\theta$ denotes the set of all parameters. Note that different from the standard cross entropy loss defined in a multi-class classification task, our JBCE loss is defined in a set of binary classification tasks.

To better meet the multi-label emotion classification task, inspired by [Zhou *et al.*, 2016], we further proposed to incorporate the prior label relations defined in the Plutchik's wheel of emotions [Plutchik, 1980] into the JBCE loss.

Plutchik's psychoevolutionary theory of emotion is one of the most influential classification approaches for general emotional responses. He considered there to be eight primary emotions: anger, fear, sadness, disgust, surprise, anticipation, trust, and joy. The wheel Plutchik's is used to illustrate different emotions in a compelling and nuanced way. It includes several typical emotions and its eight sectors indicate eight primary emotion dimensions arranged as four pairs of opposites.

In the emotion wheel, emotions sat at opposite end have an opposite relationship, while emotions next to each other are more closely related. As shown in Figure 3, we followed [Zhou *et al.*, 2016] by measuring the relations $w_{s,t}$ between the $s$-th and $t$-th emotions based on the angle between them.

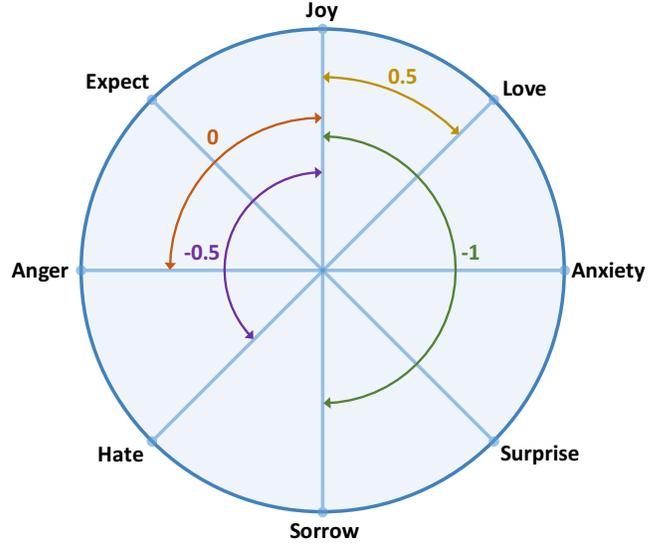

Figure 3: Plutchik's wheel of emotions.

- In case of emotion pairs with 180 degree (*i.e.*, opposite to each other), define $w_{s,t} = -1$;
- In case of emotion pairs with 90 degree, define $w_{s,t} = 0$;
- In case of emotion pairs with 45 degree, define $w_{s,t} = 0.5$;
- In case of emotion pairs with 135 degree, define $w_{s,t} = -0.5$.

On this basis, the union loss function is defined as:

$$L = -\sum_{j=1}^C \left( y_j \log p_j + (1 - y_j) \log(1 - p_j) \right) \\ + \lambda_1 \sum_{s,t} w_{s,t} (p_s - p_t)^2 + \lambda_2 ||\theta||^2. \quad (8)$$

The behind motivation is that if two emotions (such as joy and love) have a high positive correlation, we hope the prediction on the two emotions remain similar. On the contrary, if two emotions (such as joy and sorrow) have a high negative correlation, we hope the predictions on the two emotions remain different.

## 4 Experiments

### 4.1 Experimental Settings

We conduct the experiments on the Ren-CECps corpus [Quan and Ren, 2010] which was widely used in multi-label emotion classification. It contains 35,096 sentences selected from Chinese blogs. Each sentence is annotated with 8 basic emotions, such as *anger*, *anxiety*, *expect*, *hate*, *joy*, *love*, *sorrow* and *surprise*.

Table 1: Experimental results in comparison with traditional multi-label learning methods (mean±std). '↓' means 'the smaller the better'. '↑' means 'the larger the better'. Boldface highlights the best performance. '•' indicates significance difference.

| Algorithm | Ranking Loss(↓) | Hamming Loss(↓) | One-Error(↓) | Coverage(↓) | Average Precision(↑) |
|---|---|---|---|---|---|
| ECC [Read et al., 2009] | 0.3281 ± 0.0659• | 0.1812 ± 0.0940• | 0.6969 ± 0.0598• | 2.7767 ± 0.0876• | 0.5121 ± 0.0892• |
| MLLOC [Huang et al., 2012] | 0.4742 ± 0.0734• | 0.1850 ± 0.0659• | 0.6971 ± 0.0924• | 3.6994 ± 0.0764• | 0.4135 ± 0.0568• |
| ML-KNN [Zhang and Zhou, 2014] | 0.2908 ± 0.0431• | 0.2459 ± 0.0781• | 0.5339 ± 0.0954• | 2.4480 ± 0.0981• | 0.5917 ± 0.0742• |
| Rank-SVM [Zhang and Zhou, 2014] | 0.3055 ± 0.0579• | 0.2485 ± 0.0458• | 0.5603 ± 0.0921• | 2.5861 ± 0.0777• | 0.5738 ± 0.0892• |
| LIFT [Zhang and Wu, 2015] | 0.2854 ± 0.0427• | 0.1779 ± 0.0597• | 0.5131 ± 0.0666• | 2.4267 ± 0.0492• | 0.5979 ± 0.0891• |
| EDL [Zhou et al., 2016] | 0.2513 ± 0.0560• | 0.1772 ± 0.0568• | 0.5239 ± 0.0945• | 2.1412 ± 0.0235• | 0.6419 ± 0.0235• |
| JBNN (Our Approach) | **0.1511 ± 0.0030** | **0.1312 ± 0.0009** | **0.4035 ± 0.0073** | **1.7864 ± 0.0193** | **0.7171 ± 0.0041** |

Due to the inherent differences in classification problems, common metrics for multi-label classification are different from those used in single-label classification. In this study, five popular evaluation metrics are adopted in the multi-label classification experiment include Hamming Loss (HL), One-Error (OE), Coverage (Co), Ranking Loss (RL), and Average Precision (AP) [Zhang and Zhou, 2014]. Hamming loss is a label-based metric, and the rest can be divided into ranking-based metrics.

We utilize word2vec[1] to train the word vectors on the 1.1 million Chinese Weibo corpora provided by NLPCC2017[2]. The dimension of word embedding vectors is set as 200 and the size of hidden layer is set as 100. All out-of-vocabulary words are initialized to zero. The maximum sentence length is 90. All weight matrices and bias are randomly initialized by a uniform distribution $U(-0.01, 0.01)$. TensorFlow is used to implement our neural network model. In model training, learning rate is set as 0.005, $L_2$-norm regularization is set as 1e-4, the parameter $\lambda_1$ in the emotion constraint term is set as 1e-3. We use the stochastic gradient descent(SGD) algorithm and Adam update rule with shuffled mini-batch for parameter optimization.

### 4.2 Comparison with Traditional Multi-label Learning Models

In this section, we compare JBNN with six strong multi-label learning models for multi-label emotion classification, namely EDL [Zhou et al., 2016], ML-KNN [Zhang and Zhou, 2014], Rank-SVM [Zhang and Zhou, 2014], MLLOC [Huang et al., 2012], ECC [Read et al., 2009], LIFT [Zhang and Wu, 2015]. The feature space in the six compared algorithms are the same as that in [Zhou et al., 2016]. For each algorithm, ten-fold cross validation is conducted. The compared algorithms are shown as follows:

- **EDL** captures emotional relationships based on the Plutchik's wheel of emotions and incorporates them into learning algorithm to improve the accuracy of emotion detection.

- **ML-kNN** is derived from the traditional k-nearest neighbor (kNN) algorithm. The principle of maximum a posteriori (MAP) is used to determine which emotion set is related to the given sentence.

- **Rank-SVM** provides a way to control the complexity of the entire learning system with little empirical error. The Rank-SVM architecture is based on a linear model of Support Vector Machine (SVM).

- **MLLOC** (Multi-label Local Correlation) attempts to exploit emotion correlations in locally expression data. The global discrimination fitting and local correlation sensitivity are integrated into a unified framework, and solution for the optimization are proposed.

- **ECC** (Ensemble Classifier Chains) applies classifier chains, which overcomes the limitations of BR and performs better, but requires more computations and achieves high predictive performance in the ensemble framework.

- **LIFT** constructs emotion-specific features by conducting clustering analysis on its positive or negative instances, and then performs training and testing by querying the clustering results.

Table 1 shows the experimental results of the proposed method in comparison with the six strong multi-label learning methods. The two-tailed $t$-tests with 5% significance level are performed to see whether the differences between JBNN and the compared models are statistically significant. We can find that the MLLOC method is the worst, and the ECC method performs better than MLLOC. The experimental performance of MLKNN and LIFT is similar, while the performance of RankSVM is slightly worse than them. Among these traditional multi-label learning models, EDL performs the best. However, our model improves the EDL method with an impressive improvement in all kinds of evaluation metrics, i.e., 10.02% reduction in RL, 4.60% reduction in HL, 12.04% reduction in OE, 35.48% reduction in Co and 7.52% increase in AP. In short, it can be observed that our JBNN approach performs consistently the best on all evaluation measures. The improvements are all significant in all situations.

### 4.3 Comparison with Two Types of Neural Networks (BRNN and TDNN)

These models usually utilize neural networks to automatically extract features of sentence and obtain final results. In this section, we compare our proposed JBNN with two major neural networks for multi-label classification, namely BRNN and TDNN, with multi-label classification performance and computational efficiency. We implement all these approaches based on the same neural network infrastructure, use the same 200-dimensional word embeddings, and run them on the same machine. The details of implement are as follows:

- **BRNN** is implemented by constructing multiple binary

---
[1] https://code.google.com/archive/p/word2vec/
[2] http://www.aihuang.org/p/challenge.html

Table 2: Experimental results in comparison with two types of neural networks methods (mean±std). '↓' means 'the smaller the better'. '↑' means 'the larger the better'. Boldface highlights the best performance. '●' indicates significance difference.

| Algorithm | Ranking Loss(↓) | Hamming Loss(↓) | One-Error(↓) | Coverage(↓) | Average Precision(↑) |
|---|---|---|---|---|---|
| BRNN | $0.1612 \pm 0.0051$● | $0.1346 \pm 0.0015$● | $0.4243 \pm 0.0073$● | $1.8779 \pm 0.0371$● | $0.7017 \pm 0.0054$● |
| TDNN | $0.1532 \pm 0.0040$● | $0.1334 \pm 0.0013$● | $0.4148 \pm 0.0098$● | $1.7922 \pm 0.0299$ | $0.7115 \pm 0.0060$● |
| JBNN | $\mathbf{0.1511 \pm 0.0030}$ | $\mathbf{0.1312 \pm 0.0009}$ | $\mathbf{0.4035 \pm 0.0073}$ | $\mathbf{1.7864 \pm 0.0193}$ | $\mathbf{0.7171 \pm 0.0041}$ |

Table 4: The performance of JBNN and three reduced versions of JBNN (mean±std). '↓' means 'the smaller the better'. '↑' means 'the larger the better'. Boldface highlights the best performance.

| Algorithm | Ranking Loss(↓) | Hamming Loss(↓) | One-Error(↓) | Coverage(↓) | Average Precision(↑) |
|---|---|---|---|---|---|
| JBNN | $\mathbf{0.1511 \pm 0.0030}$ | $\mathbf{0.1312 \pm 0.0009}$ | $\mathbf{0.4035 \pm 0.0073}$ | $\mathbf{1.7864 \pm 0.0193}$ | $\mathbf{0.7171 \pm 0.0041}$ |
| JBNN-No-Bi | $0.1579 \pm 0.0003$ | $0.1329 \pm 0.0014$ | $0.4139 \pm 0.0051$ | $1.8413 \pm 0.0262$ | $0.7085 \pm 0.0028$ |
| JBNN-No-Att | $0.1554 \pm 0.0032$ | $0.1324 \pm 0.0016$ | $0.4117 \pm 0.0057$ | $1.8217 \pm 0.0249$ | $0.7105 \pm 0.0033$ |
| JBNN-No-EmoRel | $0.1516 \pm 0.0032$ | $0.1313 \pm 0.0012$ | $0.4046 \pm 0.0043$ | $1.7906 \pm 0.0286$ | $0.7162 \pm 0.0031$ |

Table 3: Computational Efficiency of different neural networks. Params means the number of parameters, while Time cost means runtime(seconds) of each training epoch.

| Algorithm | Params(↓) | Time Cost(↓) |
|---|---|---|
| BRNN | 2.53M | 265s |
| TDNN | 2.81M | 305s |
| JBNN | **0.28M** | **35s** |

neural networks, as shown in figure 1(a), based on Bi-LSTM and attention mechanism.

- **TDNN** is implemented using the method in [Wang *et al.*, 2016], which used a neural network based method to train one multi-class classifier and c binary classifiers to get the probability values of the c emotion labels, and then leveraged Calibrated Label Ranking (CLR) method to obtain the final emotion labels.

**Classification Performance.** In Table 2, we report the performance of JBNN, BRNN and TDNN models. From this table, we can see that our JBNN model performs significantly better than BRNN among all five kinds of evaluation metrics. Compared with the TDNN, our JBNN model is much better in Ranking Loss, Hamming Loss, One-Error, Average Precision. In general, our JBNN model performs better than both BRNN and TDNN models. The improvements according to two-tailed $t$-test are significant.

**Computational Efficiency.** We also report the size of parameters and runtime cost of BRNN, TDNN and JBNN in Table 3. From Table 3, we can find that our JBNN model is much simpler than BRNN and TDNN. For example, our JBNN model only has 0.28 M parameters, while BRNN has 2.53M parameters and TDNN has 2.81M parameters. As for runtime cost, we can see that BRNN and TDNN are indeed computationally expensive. Our JBNN model is almost 8 times faster than BRNN and 9 times faster than TDNN in model training. In summary, our JBNN model has significantly priority against BRNN and TDNN in computation efficiency.

### 4.4 Model Internal Analysis

In order to verify the effectiveness of our model and find out which part of our model contributes the most, we design the following three models based on JBNN:

- **JBNN-No-Bi** model is constructed by replacing Bi-LSTM in JBNN with LSTM.
- **JBNN-No-Att** is a simplified version of JBNN, where we remove the attention and just use the last hidden state vector as the sentence representation.
- **JBNN-No-EmoRel** utilizes only the JBCE loss function, which has not incorporated the prior label relations.

Table 4 shows the performances of all these models. From this table, we can see that JBNN-No-Bi model performs much worse than JBNN. The result verifies that Bi-LSTM may have more powerful representation capacity than LSTM. Removing the attention mechanism in JBNN results in an overall drop in performance, which proves that attention mechanism is able to capture such words that are important to the meaning of the sentence to some extent. Comparing JBNN-No-EmoRel with JBNN, we can find that incorporating the prior emotion relations leads to performance improvement for multi-label emotion classification. In summary, Bi-LSTM contributes the most to our model. This is reasonable because powerful representation has a great influence on the performance.

## 5 Conclusion

In this paper, we have proposed a joint binary neural network (JBNN) model for multi-label emotion classification. Unlike existing multi-label learning neural networks, which either needs to train a set of binary networks separately (BRNN), or although model the problem within a multi-class network, an extra threshold function is needed to transform the multi-class probabilities to multi-label outputs (JDNN), our model is an end-to-end learning framework that integrates representation learning and multi-label classification into one neural network. Our JBNN model is trained on a joint binary cross entropy (JBCE) loss. Furthermore, the label relation prior is also incorporated to capture the correlation between emotions. The experimental results show that our model is much better than both traditional multi-label emotion classification methods and the representative neural network systems (BRNN and TDNN), in both multi-class classification performance and computational efficiency.